\documentclass[letterpaper]{article} % DO NOT CHANGE THIS
\usepackage{aaai25}  % DO NOT CHANGE THIS
\usepackage{times}  % DO NOT CHANGE THIS
\usepackage{helvet}  % DO NOT CHANGE THIS
\usepackage{courier}  % DO NOT CHANGE THIS
\usepackage[hyphens]{url}  % DO NOT CHANGE THIS
\usepackage{graphicx} % DO NOT CHANGE THIS
\usepackage{natbib}  % DO NOT CHANGE THIS AND DO NOT ADD ANY OPTIONS TO IT
\usepackage{caption} % DO NOT CHANGE THIS AND DO NOT ADD ANY OPTIONS TO IT
\usepackage{algorithm}
\usepackage{algorithmic}

\usepackage{newfloat}
\usepackage{listings}
\DeclareCaptionStyle{ruled}{labelfont=normalfont,labelsep=colon,strut=off} % DO NOT CHANGE THIS
\floatstyle{ruled}
\newfloat{listing}{tb}{lst}{}
\floatname{listing}{Listing}
\usepackage[utf8]{inputenc} % allow utf-8 input
\usepackage[T1]{fontenc}    % use 8-bit T1 fonts
\usepackage{url}            % simple URL typesetting
\usepackage{booktabs}       % professional-quality tables
\usepackage{amsfonts}       % blackboard math symbols
\usepackage{nicefrac}       % compact symbols for 1/2, etc.
\usepackage{microtype}      % microtypography
\usepackage{xcolor}         % colors
\usepackage{graphicx}
\usepackage{subfigure}
\usepackage{amsmath}
\usepackage{amssymb}
\usepackage{multirow}
\usepackage{bbding}
\usepackage{amsmath}
\usepackage{svg}
\usepackage{algorithm}% http://ctan.org/pkg/algorithms
\usepackage{algorithmic}
\usepackage{caption}
\usepackage{textcomp}
\usepackage{dblfloatfix}
\providecommand{\eg}{\emph{e.g.}}
\providecommand{\ie}{\emph{i.e.}}

\title{AsyncDSB: Schedule-Asynchronous Diffusion Schr\"odinger \\ Bridge for Image Inpainting}
\author{
    Zihao Han$^{1}$, Baoquan Zhang$^{1}$\thanks{Corresponding author.}, Lisai Zhang$^{4}$, Shanshan Feng$^{3}$, Kenghong Lin$^{1}$, Guotao Liang$^{1}$, Yunming Ye$^{1}$, Xiaochen Qi$^{2}$, Guangming Ye$^{2}$
}
\affiliations{
    \textsuperscript{\rm 1} Harbin Institute of Technology, Shenzhen;
    \textsuperscript{\rm 2} ShenZhen SiFar Co., Ltd.;
    \textsuperscript{\rm 3} Centre for Frontier AI Research;
    \textsuperscript{\rm 4} Shengshu AI.\\
    
    \{200110201, linkenghong\}@stu.hit.edu.cn, \{baoquanzhang, victor\_fengss, yunming ye\}@hit.edu.cn, LisaiZhang@foxmail.com, lianggt@pcl.ac.cn, \{joeqxc1974, kolaygm\}@gmail.com
}

\usepackage{bibentry}
\begin{document}

\maketitle

\begin{abstract}
%背景
  % Image inpainting is a challenging task, which aims to restore an corrupted image with few visible
  % context. Recently, diffusion bridge-based methods effectively address this challenge by modeling the translation between corrupted and target images in a pixel-synchronous Schr\"odinger diffusion bridge manner.
  
  %Image inpainting是最近的桥扩散方法的重要任务
  %Image inpainting is an important image generation task and has been the prominent task of diffusion bridge-based methods. 
  Image inpainting is an important image generation task, which aims to restore corrupted image from partial visible area. 
  % 描述一下，引出synchronous
  Recently, diffusion Schr\"odinger bridge methods effectively tackle this task by modeling the translation between corrupted and target images as a diffusion Schr\"odinger bridge process along a noising schedule path. Although these methods have shown superior performance, in this paper, we find that 1) existing methods suffer from a schedule-restoration mismatching issue, \ie, the theoretical schedule and practical restoration processes usually exist  a large discrepancy, which theoretically results in the schedule not fully leveraged for restoring images; and 2) the key reason causing such issue is that the restoration process of all pixels are actually asynchronous but existing methods set a synchronous noise schedule to them, \ie, all pixels shares the same noise schedule. %The assumption behind such idea is that all pixels are independent of each other (\ie, no temporal interdependence) in the generation sequence.
  %动机 
  %However, in this paper, we conduct an in-depth analysis on the diffusion bridge process to show that the translation of different pixels are actually in different speed, \ie, the high-frequency structure parts are restored early than the low-frequency details pixels,
  %which means the schedule of existing Schr\"odinger diffusion bridge actually does not fit the translation well and limits its performance.
  % , which could be better modeled with an asynchronous schedule. 
  % This means that such schedule-synchronous Schr\"odinger diffusion bridge manner is not optimal, which limits the performance. 
%方法&贡献
  %However, in fact, the generation process of all pixels are not independent but exists a temporal interdependence, \ie, some pixels containing outline and texture information (\ie, high frequency component) are usually restored first and then some pixels with lower frequency component are repaired in later.
  To this end, we propose a schedule-\textbf{Async}hronous \textbf{D}iffusion \textbf{S}chr\"odinger \textbf{B}ridge (\textbf{AsyncDSB}) for image inpainting. Our insight is preferentially scheduling pixels with high frequency (\ie, large gradients) and then low frequency (\ie, small gradients). 
  % Based on this, given a corrupted image, we first restore its gradient map in corrupted area through an adversarial learning manner. Then, we regard the image gradient as prior and design a simple yet effective pixel-asynchronous noise schedule strategy to enhance the diffusion Schr\"odinger bridge. 
  Based on this insight, given a corrupted image, we first train a network to predict its gradient map in corrupted area. Then, we regard the predicted image gradient as prior and design a simple yet effective pixel-asynchronous noise schedule strategy to enhance the diffusion Schr\"odinger bridge.
  Thanks to the asynchronous schedule at pixels, the temporal interdependence of restoration process between pixels can be fully characterized for high-quality image inpainting. Experiments on real-world datasets show that our AsyncDSB achieves superior performance, especially on FID with around 3\% $\sim$ 14\% improvement over state-of-the-art baseline methods.

\end{abstract}


\section{Introduction}

%Image inpainting  (also known as image completion) is a long-standing low-level computer vision task, which aims to restore the corrupted image from the content of the visible area. 
%Early studies mainly focus on non-learning inpainting strategy to address this above challenge, such as diffusion [2], PatchMatch [1], etc., \ie, using statistical information of the visible area to infer the missing details. These methods can produce inpainting results with reasonable structure and texture when the mask area is small, or the undegraded part shows well-defined geometry. However, these methods often perform poorly when being required to recover semantics since they lack image semantic priors.

% 介绍image inpainting 任务的价值，并交待简单背景（什么问题，早期状态）
Image inpainting, also known as image completion, is a well-established low-level computer vision task. Its objective is to restore a corrupted image by leveraging information from the visible portion. Early studies primarily focus on non-learning inpainting methods to tackle this challenge, such as diffusion \cite{ballester2001filling, bertalmio2000image, zomet2003learning} and PatchMatch \cite{barnes2009patchmatch, bertalmio2003simultaneous, criminisi2003object, criminisi2004region}. These methods infer the missing details in corrupted area by using statistical information from visible area, and achieve reasonable structure and texture for small region or the ungraded parts with well-defined geometry. However, these methods often struggle to recover large region effectively due to the lack of semantic inference ability.

\begin{figure*}
    \centering
    %\resizebox{0.3\linewidth}{!}{\includegraphics[]{fig/noise-schedule2.pdf}}
    \subfigure[Noise schedule ($\beta_t$)]{
        \begin{minipage}[t]{0.225\textwidth}
            \centering
            \includegraphics[width=0.98\columnwidth]{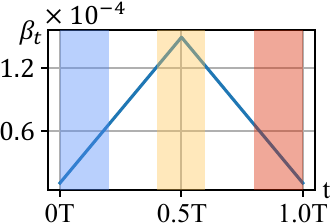}
            
        \end{minipage}%
        \label{fig_1a}
    }%
    % \quad
    \subfigure[The Restoration Process of I$^2$SB\cite{liu2023i2sb} ($T$ is the number of denoising steps.)]{
        \begin{minipage}[t]{0.755\textwidth}
            \centering
            \includegraphics[width=0.98\columnwidth]{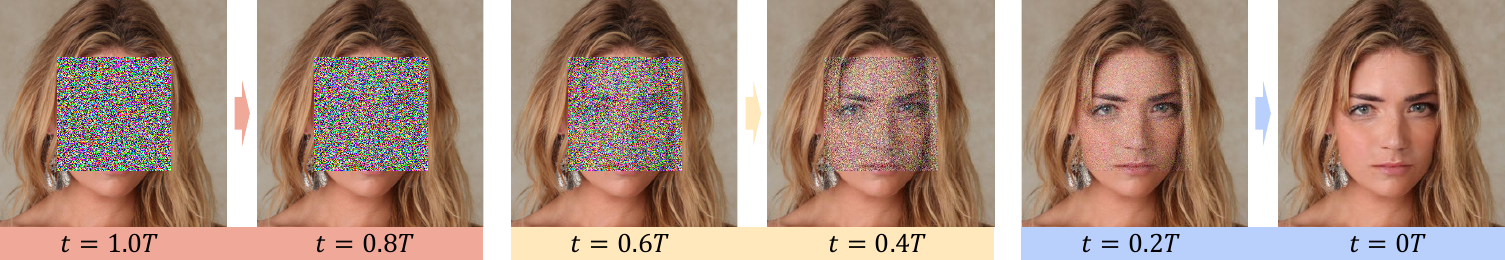}
        \end{minipage}%
        \label{fig_1b}
    }%
    \caption{Connection between noise schedule ($\beta_t$) and the restoration process of I$^2$SB\cite{liu2023i2sb}. }
    \label{fig_1}
\end{figure*}

% 深度学习方法的简单脉络，突出其局限性，并介绍薛定谔桥（包含其功能，另外也提一下其schedule的设计，埋下伏笔）
To address the above challenge, deep learning inpainting methods are proposed and have received wide attention recently. These methods can be roughly divided into three categories, \ie, generative adversarial networks (GAN) -based \cite{dong2022incremental, li2022misf}, diffusion models-based \cite{liu2024structure, lugmayr2022repaint, xia2023diffir}, and diffusion bridge models-based \cite{li2023bbdm, liu2023i2sb} approaches. Among them, the diffusion bridge models-based approaches \cite{li2023bbdm, liu2023i2sb} have shown remarkable success due to its realistic and robust generation ability, especially diffusion Schr\"odinger bridge methods (a typical method is I$^2$SB \cite{liu2023i2sb}, which models the translation between corrupted and target images as a diffusion Schr\"odinger bridge and then learns it in a diffusion training manner.) 
Although I$^2$SB has shown superior performance on image inpainting, we notice that their noise schedule $\beta_t$ is only roughly set as a symmetric form (see Figure~\ref{fig_1a}) with experience setting for all images and pixels. However, there is few works to analyze the working mechanism of such noise schedule and argue the rationality of such experience setting. 

%existing I$^2$SB still employ the widely used traditional diffusion noise schedule \cite{ddpm}, \ie, applying the same $\beta_t$ for all the pixels. 
%The assumption behind such design for traditional diffusion process is that the noise on all pixels are independent, \ie, there is no temporal dependence on all pixels.
% However, there is no such restriction for Schr\"odinger bridge. 
%Therefore, there is a nature but important question: \emph{Is directly applying previous noise schedule to diffusion Schr\"odinger bridge reasonable?}
%Two nature questions are \emph{1) What role does the noise schedule play in the diffusion Schr\"odinger bridge? and 2) Is directly applying previous noise schedule to diffusion Schr\"odinger bridge reasonable?}
%Although existing DSB methods have shown superior performance, we find that all methods focus on experience manner to set the noise schedule and few existing works analyze working mechanism of such design and argue its rationality. %In particular, a representative diffusion bridge-based method is diffusion Schr\"odinger bridge methods \cite{liu2023i2sb} (\ie, DSB), which model the translation between corrupted and target image as a nonlinear diffusion Schr\"odinger bridge and then restore the corrupted images along a noise schedule path.
%Recently, some studies attempt to further improve the restrorement performance of DSB but most works focus on simplifying the theory of Schr\"odinger bridge or improving its slove efficiency. To our knowledge, 

%Intuitively, the generation process of all pixels are not independent but exists a temporal interdependence, \ie, some pixels containing outline and texture information (\ie, high frequency component) are usually restored first and then some pixels with lower frequency component are repaired in later. 
To this end, in this paper, we conduct an in-depth theoretical and qualitative analysis on the noise schedule $\beta_t$ and find that: 
 \textbf{1)} the noise schedule $\beta_t$ actually controls the restoration speed of corrupted region (see theoretical analysis of Section~\ref{sec_3_3}). As shown in Figure~\ref{fig_1}, during the noise schedule, the main contents of an image are restored in few schedule steps between $t=0.6T\rightarrow 0.4T$ since they have a large $\beta_t$ value. Instead, the other steps close to $0T$ and $1.0T$ are only responsible for cold start and detail refinement. It is such three-stage rough-fine restoration schedule that makes existing I$^2$SB method \cite{liu2023i2sb} achieve superior performance on image inpainting;
 %which means the restoration speed is typically non-linear and only few intermediate schedule steps (see the schedule at $t=0.6T\rightarrow 0.4T$ of Figure~\ref{fig:noise-schedule}) are used for restoring each pixels; 
\textbf{2)} However, the I$^2$SB simply set the same schedule $\beta_t$ for all pixels. The assumption behind such design is the three-stage restoration process of all pixels are pixel-synchronous. Unfortunately, the restoration process of pixels are not synchronous but actually asynchronous. For example, in general, the high-frequency structure pixels (\eg, face contour) are restored early than the low-frequency details pixels (\eg, face colors). 
% This results in the setting mismatch between 
Therefore, the synchronous noise schedule actually mismatch the asynchronous pixels restoration process;
and \textbf{3)} Because of the above setting mismatch, the existing I$^2$SB method \cite{liu2023i2sb} suffers from a schedule-restoration mismatching issue, \ie, there is usually a large discrepancy between the theoretical denoise schedule and practical restoration process (see our insight of Section~\ref{sec_3_3} for more details). %and 
%\textbf{3)} the key reason causing such issue is that the restoration process of all pixels are actually asynchronous (as shown in Figure~\ref{}, the low-frequency structure parts like face contour are restored early than the low-frequency details pixels) but existing methods set all pixels to a synchronous noise schedule, \ie, all pixels shares the same noise schedule. This fact means that existing synchronous schedule of Schr\"odinger diffusion bridge actually is not optimal, which limits its inpainting performance.
%; and 2) these corrupted pixels in the image are not restored in a synchronous manner but exists a prior order. For example, as shown in Figure~\ref{}, the low-frequency structure parts (\eg, face contour and organ shape) are restored early than the low-frequency details pixels (\eg, face colors). 
These facts mean that existing pixel-synchronous schedule used in existing I$^2$SB method \cite{liu2023i2sb} is actually not optimal, which limits its image inpainting performance.

To address this problem, we present a novel schedule-Asynchronous Schr\"odinger Diffusion Bridge (AsyncDSB) framework for image inpainting. The main idea is that introducing a pixel-asynchronous noise schedule for each pixel of corrupted images such that the sequential dependence of pixel restoration can be fully characterized. Intuitively, the pixels with high frequency information (\ie, large gradients) are restored early than low-frequency pixels (\ie, small gradients). Based on this insight, given a corrupted image, we first restore its gradient map (\ie, frequency component) through an adversarial learning manner. Then, we regard the image gradient as frequency priors and design a simple yet effective pixel-asynchronous noise schedule strategy for diffusion Schr\"odinger bridge. Thanks to the asynchronous schedule at pixels, the temporal interdependence between pixel restoration can be fully captured for high-quality image inpainting. 
%we first design an image gradient adversarial learning to generate structure information (\ie, high frequency profile information) for corrupted images. Then, we regard the structure information as prior, and design a simple yet effective structure-guided asynchronous noise schduling strategy for Schr\"odinger diffusion bridge. Finally, we perform image inpainting in an asynchronous Schr\"odinger diffusion bridge manner. The advantage of such design is the dependency relationship between pixeles can be fully characterized for high-quality inpainting. 

 %贡献：
 % 发现现在Schr\"odinger桥的问题，该问题限制了该sota方法的能力or在xx方面的表现；（这个发现的价值）
 % 提出一种schedule-asyn 桥，具备更强的扩散灵活性（更灵活的schedule上限应该更高）；
 % 方法的效果好，无需训练就能带来提升；
 % 其他：训练速度xx
In summary, our contributions are three-fold as follows:

\begin{itemize}
\item We inspect the working mechanism of existing noise schedule $\beta_t$ and find that applying the same noise schedule to all pixels results in a schedule-restoration mismatching issue, \ie, there is usually a large discrepancy between the theoretical noise schedule and practical restoration processes. This limits the performance of existing methods.

\item To address the above issue, we propose a novel schedule-Asynchronous Schr\"odinger Diffusion Bridge (AsyncDSB) framework. It effectively aligns the theoretical schedule and practical restoration process such that better inpainting performance can be achieved. 

\item Extensive experiments conducted on two datasets (\ie, CelebA-HQ and Places2) demonstrate the superior performance of the proposed AsyncDSB method in comparison to various state-of-the-art image inpainting methods.

\end{itemize}  

\section{Related Work}
\subsection{Image Inpainting}
Image inpainting is a challenging task, which aims restore a corrupted image with partial visible area. %Early studies primarily focus on non-learning inpainting methods to tackle this challenge, which aims to complete the corrupted region in an information diffusion \cite{ballester2001filling, zomet2003learning, bertalmio2000image} or patch matching \cite{ bertalmio2003simultaneous, criminisi2003object, criminisi2004region, barnes2009patchmatch} manner of visible area. The assumption of such methods is that corrupted image holes share similar content to visible regions, which means that these methods only perform well for small region but cannot hallucinate unique semantic content. To address this issue, recent studies turns to deep learning methods, which can be roughly categorised into three groups: 
Recently, deep learning-based methods are proposed to tackle this challenge, which mainly contains three groups: 
(1) \emph{GAN-based approaches}\cite{bendel2024regularized, chu2023rethinking,  feng2022cross, li2022mat,nazeri2019edgeconnect, sargsyan2023mi, wang2022dual, zeng2021cr}. This type of methods focuses on the encoder-decoder network to restore corrupted image and learn it through adversarial learning manner, such as MISF \cite{li2022misf} and ZITS \cite{dong2022incremental}. %For example, in \cite{li2022misf}, Li et al. propose a multi-level interactive filtering-based adversarial learning approach, which explores abundant image-level and semantic-level semantic relations for high-fidelity image inpainting; 
(2) \emph{Diffusion models-based approaches}\cite{liu2024structure, lugmayr2022repaint, xia2023diffir, zhang2023towards}. This group of methods aims to design a denoising diffusion probabilistic model with input-conditional UNet to transform image inpainting as a conditional generation task, such as StrDiffusion \cite{liu2024structure}, Repaint \cite{lugmayr2022repaint}, DiffIR \cite{xia2023diffir}. %For example, in \cite{xia2023diffir}, Xia et al. propose an efficient diffusion model that utilizes a compact prior and dynamic transformer for image restoration, achieving state-of-the-art performance with reduced computational cost. 
(3) \emph{Diffusion bridge models-based approaches}\cite{de2021diffusion, li2023bbdm, liu2023i2sb, liu2023learning, peluchetti2023diffusion, shi2024diffusion, su2022dual}.  This kind of approach is a new model of conditional diffusion, which directly models the translation between corrupted image and target image in a diffusion bridge manner \cite{li2023bbdm, liu2023i2sb}. For example, in \cite{liu2023i2sb}, Liu et al. propose a nonlinear diffusion bridge, \ie, diffusion Schr\"odinger bridge (called I$^2$SB), to model the translation between corrupted and target images. 
The non-linear characteristic of diffusion Schr\"odinger bridge significantly improves image inpainting performance. %in \cite{li2023bbdm}, Li et al. models the translation as a stochastic Brownian bridge process, learning the translation between two domains directly through a bidirectional diffusion process. Instead of using Brownian bridge, Li et al. 

In this paper, we focus on I$^2$SB due to its superiority but find that 1) it suffers from a schedule-restoration mismatching issue (see Section~\ref{sec_3_3}); and 2) its reason is pixel-synchronous noise schedule setting. To this end, we present a novel pixel-asynchronous schedule strategy, which effectively aligns the theoretical schedule and
practical restoration process for better inpainting performance.

%In this paper, we mainly focus on diffusion  Schr\"odinger bridge-based approaches due its superiority. In particular, we  find that existing methods all focus on sharing the same noise schedule for all pixels, which results in a schedule-restoration mismatching issue, \ie, the theoretical noise schedule and practical restoration processes exist usually a large gap. This limits the inpainting performance of existing methods. To this end, we propose a novel pixel-asynchronous schedule strategy for enhancing diffusion Schr\"odinger bridge. It effectively aligns the theoretical denoise schedule and practical restoration process such that better inpainting performance can be achieved.

\subsection{Noise Schedule in Diffusion Models}
Noise schedule $\beta_t$ is an important hyper-parameter in diffusion models, which determines how much weight we assign to each denoising step. As existing noise schedules (\eg, linear-based \cite{ho2020denoising} and cosine-based \cite{nichol2021improved} schedule) are only empirical settings, some studies raised a questionable point of view on noise schedule recently and propose some novel noise schedules for diffusion models, such as re-scaling schedule \cite{lin2024common}, WSNR-equivalent schedule \cite{guo2023rethinking}, compound schedule \cite{chen2023importance}, and optimization-based schedule \cite{strasman2024analysis}. %For example,  in \cite{lin2024common}, Lin et al. design a novel schedules rescaling strategy  to enforce zero signal-to-noise ratio (SNR) for noise schedules. 
However, to our best knowledge, there is few works to explore it for diffusion bridges, especially I$^2$SB. In this paper, we take an in-depth analysis on noise schedule and propose a novel pixel-asynchronous schedule, which effectively improves the inpainting performance of I$^2$SB.

\section{Methodology}
\subsection{Problem Definition}
Formally, given a target image $x_g$, we assume that the target image $x_g$ is corrupted to a corrupted image $x_c$ along an image mask $x_m$ where $x_m^{i,j}=1$ if pixel $x_c^{i,j}$ at spatial position $(i,j)$ is corrupted region otherwise $x_m^{i,j}=0$, \ie, $x_c = (1-x_m) \times x_g$. The goal of image inpainting is to restore the missing area (\ie, $x_m \times x_g$) within the corrupted image $x_c$. In particular, the restored areas should blend seamlessly with the surrounding visible region and be logically consistent.

\subsection{Preliminaries: Diffusion Schr\"odinger Bridge for Image Inpainting}
\label{sec3_2}
Recently, diffusion Schr\"odinger bridge models (DSB), as a new  conditional diffusion models, has achieve success on image inpainting. Different from previous input-conditional diffusion models, DSB model directly learns the transformation processes between target image distributions $p_{x_g}$ and corrupted image distribution $p_{x_c}$ via a nonlinear Schr\"odinger bridge. Its key challenge is how learn the Schr\"odinger bridge on image inpainting. Next, we briefly review the Schr\"odinger bridge and its adaptation on image inpainting, \ie, Diffusion Schr\"odinger Bridge \cite{liu2023i2sb}.

%\textbf{\emph{Schr\"odinger Bridge}} is a nonlinear score-based model which defines an optimal transport between two arbitrary distributions $p(x_0)$ and $p(x_1)$. Here, the  variable $x(t)$ at each time step $t \in [0,1]$ can be characterized as the following forward and backward stochastic differential equations (SDEs):  
\textbf{\emph{Schr\"odinger Bridge (SB)}} is an entropy-regularized optimal transport problem, which has been widely applied to quantum mechanics \cite{pavon2002quantum} and optimal transport and control \cite{chen2016relation}. Given two arbitrary distributions $p(x_0)$ and $p(x_1)$, SB aims to seek an optimal path pair between $p(x_0)$ and $p(x_1)$  with the following forward and backward stochastic differential equations (SDEs):  
\begin{eqnarray}
    \resizebox{.92\hsize}{!}{$
    \begin{aligned}
        &\textbf{Forward}:\ d x_t = [f_t + \beta_t \nabla \log {\Psi}(x_t, t)] d t + \sqrt{\beta_t} d W_t, \ t \in [0, 1],\\
        &\textbf{Backward}:\ d x_t = [f_t - \beta_t \nabla \log \hat{\Psi}(x_t, t)] d t + \sqrt{\beta_t} d \hat{W}_t, \ t \in [0, 1],
    \end{aligned}
    $}
    \label{eq0}%
\end{eqnarray}
% \begin{equation}
%         d X_t = [f_t - \beta_t \nabla log \hat{\Psi}(X_t, t)] d t + \sqrt{\beta_t}d W_t
%        % \rd X_t &= [f_t - \beta_t~\gradlog \Psihat(X_t, t) ] \dt + \sbeta\rd \barW_t
%     \label{eq:sb-sde}%
% \end{equation}
where the forward and backward SDEs denote the optimal transformation from $t=0$ to $t=1$ and its reverse from $t=1$ to $t=0$, respectively; $x_0 \sim p_{x_0}$ and $x_1 \sim p_{x_1}$ denote samples from two arbitrary distributions; $W_t$ and $\hat{W}_t$ denote the standard Brownian motion and its reversed counterpart, respectively; $f_t$ is a drift term; and $\Psi(\cdot)$ and $\hat{\Psi}(\cdot)$ denotes the time-varying energy functions.
% with following PDEs:
% \begin{equations}
% \begin{align}
%    & \begin{cases}
%   \frac{\partial \Psi(x,t)}{\partial t}    = - \nabla \Psi^\top f - \frac{1}{2} \beta \Delta \Psi \\[3pt]
%   \frac{\partial \hat{\Psi}(x,t)}{\partial t} = - \nabla \cdot (\hat{\Psi} f) + \frac{1}{2} \beta \Delta \hat{\Psi}
%   \end{cases} 
%   \\
%    \text{s.t. }
%   \Psi(x,0) &  \hat{\Psi}(x,0){=}p_\mathcal{A}(x),
%       \Psi(x,1) \hat{\Psi}(x,1){=}p_\mathcal{B}(x) 
% \end{aligne}
% \label{eq:sb-pde3}
% \end{equations}
% Finally, the optimal transformation can be obtained by solving the above SDEs defined in Eq~\eqref{eq:sb-sde}.

\emph{\textbf{Diffusion Schr\"odinger Bridge}}
In fact, the image inpainting process can also be regarded as a Schr\"odinger bridge when we set the target image distributions $p_{x_g}$ and its corrupted image distributions $p_{x_c}$ as $p_{x_0}$ and $p_{x_1}$, respectively, \ie, $p_{x_0}=p_{x_g}$ and $p_{x_1}=p_{x_c}$. However, due to the complexity of SB, the key challenge of adapting SB to image inpainting is how to solve the SDEs defined in Eq~\eqref{eq0}. To address this challenge, Liu et al. \cite{liu2023i2sb} propose a tractable diffusion-based Schr\"odinger bridge (called I$^2$SB \cite{liu2023i2sb}) recently, which simplifies the SDEs of Eq~\eqref{eq0} by dropping $f := 0$ and deduce an analytic solution for forward SDEs with the following Gaussian posterior: 
\begin{equation}
    \resizebox{.9\hsize}{!}{$
    \begin{aligned}
        &x_t \sim q(x_t|x_0, x_1) = N(x_t; \mu_t, \Sigma_t), \\ 
        &\mu_t = \frac{\overline{\sigma}_t^2}{\overline{\sigma}_t^2 + \sigma^2_t} x_0 +
              \frac{\sigma^2_t    }{\overline{\sigma}_t^2 + \sigma^2_t} x_1,
        \Sigma_t = \frac{\sigma_t^2 \overline{\sigma}_t^2}{\overline{\sigma}_t^2 + \sigma^2_t} \cdot I,
    \end{aligned}
    $}
\label{eq2}
\end{equation}
where $\sigma_t$ and $\overline{\sigma}_t$ are two variances accumulated from either sides along a noise schedule $\beta_t$, \ie,  
\begin{equation}
\sigma_t^2 = \int_{0}^{t}\beta_\tau d\tau,\ \overline{\sigma}_t^2 = \int_{t}^{1}\beta_\tau d\tau ,
\label{eq3}
\end{equation} where the noise schedule $\beta_\tau$ is experimentally set to a simple symmetric form (see Figure~\ref{fig_1a}). Based on the above analytic solution, the score function for $\nabla log {\hat{\Psi}}(x_t, t)$ can be directly calculated, \ie, $\nabla log {\hat{\Psi}}(x_t, t) = \frac{x_t-x_0}{\sigma_t^2}$, which is further used for learning a score function $s_{\theta}(x_t,t)$. Following standard diffusion (\ie, DDPM \cite{ho2020denoising}), the overall score-matching objective $L$ can be defined as:
\begin{equation}
 L = \| s_{\theta}(x_t, t) - \frac{x_t-x_0}{\sigma_t}\|.
 \label{eq4}
\end{equation}
After training (\ie, during inference), 
%the score function $s_{\theta}(x_t,t)$ can be directly used to estimate the target image $\hat{x}_0$, \ie, $\hat{x}_0=x_t-s_{\theta}(x_t, t)\sigma_t$. Finally, 
given a corrupted image, \ie, $x_1=x_c$,  we can directly traverse the restoration process induced by Eq~\eqref{eq0} by the following recursive posterior sampling in DDPM:
\begin{equation}
    \resizebox{.9\hsize}{!}{$
    \begin{aligned}
	% q(x_n|\hat{x}_0,x_N) = \int.
        q(x_{t-\frac{1}{T}}|x_{t}) = 
        {\cal N} \left (x_{t}-\int_{t-\frac{1}{T}}^{t}\left [ -\beta_t \nabla log \hat{\Psi}(x_{t}, t) \right ]dt, \int_{t-\frac{1}{T}}^{t} \beta_t dt \right ) , 
        t=1,1-\frac{1}{T},...,0,
    \end{aligned}
    $}
\label{eq:ddpm}
\end{equation} where $\nabla log \hat{\Psi}(x_t, t) = \frac{s_{\theta}(x_t, t)}{\sigma_{t}}$ and $T$ denotes the total time step of discretized SDE. Following \cite{liu2023i2sb}, $T=1000$ is used. Finally, a restored image, \ie, $\hat{x_g}=\hat{x_0}$, can be achieved at time step $t=0$.

\subsection{A Closer Look at Existing Noise Schedule $\beta_t$}
\label{sec_3_3}
Although existing I$^2$SB \cite{liu2023i2sb} has shown superior performance, we find that 1) existing method focuses on experience manner to set the noise schedule $\beta_t$ defined in Eq~\eqref{eq3}; and 2) as shown in Figure~\ref{fig_1a}, the noise schedule $\beta_t$ is usually set as a symmetric form for all images and pixels. However, few works analyze the working mechanism of such design and argue the rationality of such experience setting. 
In this paper, we conduct an in-depth theoretical and qualitative analysis to clarify this point.
%Note that in existing SDB methods \cite{}, as shown in Figure~\ref{fig:noise-schedule}, $\beta_\tau$ is only set empirically to a symmetric noise schedule for all images even pixels. However, to our knowledge, few works analyze its working mechanism and argue its rationality. In this paper, we attempt to clarify this question. %However, a natural question is \emph{Is such symmetric noise schedule is optimal?} Our answer is no, for reasons that will be discussed in Section~\ref{sec_3_3}.   

%As shown in Section 3.2, existing methods mainly focus on the symmetric noise schedule to model the transformation process between the target and corrupted image distribution, which is a typical experience setting. In this section, we first analyze its basic principle and then clarify its rationality. %However, a intuitive question is \emph{Is such an experience noise schedule strategy necessarily optimal?} To answer this question, we first conduct a qualitative and theoretical analysis on noise schedule, and then attempt to answer this question.

\paragraph{Theoretical Analysis on Noise Schedule} 
\label{sec_3_3_1}
Intuitively, the noise schedule $\beta_t$ may control the restoration process (\ie, speed) of image inpainting along the diffusion Schr\"odinger bridge. To verify this point, in this part, we conduct a theoretical analysis on the transformation path of diffusion Schr\"odinger bridge. Formally, given a pair corrupted and target samples $(x_c, x_g)$. Then, we set the pair sample as the two endpoints of SB, \ie, $x_0=x_g$ and $x_1=x_c$. After that, following the Eq~\eqref{eq2}, we can represent each restored image $x_t$ at time step $t$ on the transport path as:
\begin{equation}
x_{t}=\frac{\bar{\sigma}_t^2}{\bar{\sigma}_t^2 + \sigma_t^2}x_0 + \frac{\sigma_t^2}{\bar{\sigma}_t^2 + \sigma_t^2}x_1+\frac{\bar{\sigma}_t\sigma_t}{\sqrt{\bar{\sigma}_t^2 + \sigma_t^2}}z,
\label{eq4}
\end{equation} 
where $z$ denotes a Gaussian noise, \ie, $z \sim \mathcal{N}(0,1)$. Then, the restoration speed of image inpainting process at time $t$ can be characterized as a derivative of $x_t$ about the time $t$. That is,
\begin{equation}
    \resizebox{.9\hsize}{!}{$
    \begin{aligned}
        \frac{\partial x_t}{\partial t}~&=\frac{\partial\frac{\bar{\sigma}_t^2}{\bar{\sigma}_t^2 + \sigma_t^2}}{\partial t}x_0 + \frac{\partial\frac{\sigma_t^2}{\bar{\sigma}_t^2 + \sigma_t^2}}{\partial t}x_1 + \underbrace{\frac{\partial \frac{\bar{\sigma}_t\sigma_t}{\sqrt{\bar{\sigma}_t^2 + \sigma_t^2}}}{\partial t}z}_{Term 3}  \\
        ~&=\frac{\sigma_t^2\frac{\partial \bar{\sigma}_t^2}{\partial t}-\frac{\partial \sigma_t^2}{\partial t}\bar{\sigma}_t^2}{(\bar{\sigma}_t^2+\sigma_t^2)^2}x_0 + \frac{ \bar{\sigma}_t^2\frac{\partial \sigma_t^2}{\partial t}-\frac{\partial \bar{\sigma}_t^2}{\partial t}\sigma_t^2 }{(\bar{\sigma}_t^2+\sigma_t^2)^2}x_1\\
        ~&=\frac{(x_1-x_0)(\beta_t\bar{\sigma}_t^2+\sigma_t^2\beta_t)}{(\bar{\sigma}_t^2+\sigma_t^2)^2} 
        ~=\beta_t\frac{(x_1-x_0)}{\bar{\sigma}_t^2+\sigma_t^2}, 
    \end{aligned}
    $}
\label{eq5}
\end{equation} 
where $Term 3$ can be overlooked since its expectation is zero. In particular, the corrupted image $x_1$ satisfies $x_1 = x_0 \times (1-x_m)$, which suggests that the derivative $\frac{\partial x_t}{\partial t}$ can be further expressed as: 
\begin{equation}
\frac{\partial x_t}{\partial t}=\beta_t\frac{x_0 \times (1-x_m)-x_0}{\bar{\sigma}_t^2+\sigma_t^2} = -\beta_t \frac{x_0 x_m}{\bar{\sigma}_t^2+\sigma_t^2}.
\label{eq6}
\end{equation}Note that 1) $x_m$ denotes the mask of corrupted region, \ie, $x_m^{i,j}=1$ if pixel $x_1^{i,j}$ at spatial position $(i,j)$ is corrupted region otherwise $x_m=0$, and 2) $\bar{\sigma}_t^2+\sigma_t^2= \int_{0}^{1}\beta_\tau d\tau$ is actually a constant and can be overlooked. Let's take a closer look at the Eq~\eqref{eq6}, we can see that the restoration speed $\frac{\partial x_t}{\partial t}$ of image inpainting process is only relevant with the noise schedule $\beta_t$ and the value $x_0^{ij}$ of each pixel $(i,j)$ at the corrupted region. In particular, the noise schedule $\beta_t$ controls the overall restoration tread and the pixel value $x_0^{ij}$ controls the amplitude of restoration at corrupted region. 

% To more clearly illustrate this point, in Figure~\ref{}, we randomly select an image from CelebA-HQ test set, and then visualize the restorment speed $\frac{\partial x_t^{ij}}{\partial t}$ of some plxels $x_g^{ij}$. From Figure~\ref{}, we can find that 1) the overall tread are very consist for all pixels which is reasonable since all pixels share the same noise schedule; however 2) the only difference is the amp, \ie, big pixel value own biger amp. 

Based on this fact, let's revisit the the noise schedule $\beta_t$ shown in Figure~\ref{fig_1a}, we can see that $\beta_t$ is  symmetric and two bound points are small but middle point is big. This means that the restoration process at two bound points are very slow (a reasonable explain is that the early slow speed is to achieve a good cold start, however the later is to achieve detail refinement) and only few intermediate schedule steps are used for quickly restoring images. It is such nonlinear noise schedule with well cold start and detail refinement that I$^2$SB achieves superior performance on image inpainting.

\paragraph{Our Insight: Schedule-restoration Mismatching Issue}  The Eq~\eqref{eq6} is only theoretical analysis result, a nature question is \emph{Is the practical restoration process really like theoretical schedule?}. To answer this question, we evaluate the SSIM performance of CelebA-HQ test set at each denoising time step $t$ from $t=1.0T$ to $t=0T$ ($T$=1000) and calculate its derivative. For clarity, we normalize its amplitude for removing the impact of pixel values.

\begin{figure}
\centering
\includegraphics[width=0.58\hsize]{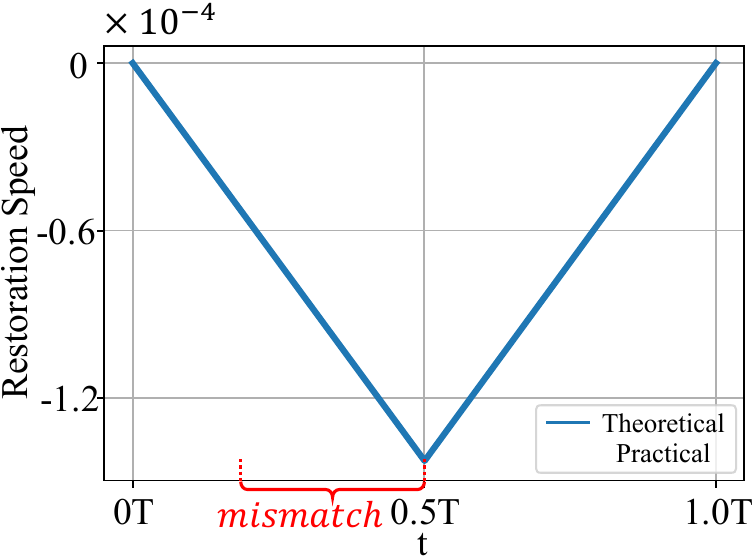}
\caption{Qualitative Analysis.}
\label{fig2}
\end{figure}
% \centering
% \includegraphics[width=0.58\hsize]{fig/restoration_speed.pdf}
% \label{fig2b}
% \captionof{figure}{Qualitative Analysis.}
% \label{fig2}

The result is shown in Figure~\ref{fig2}. From Figure~\ref{fig2}, we can see that 1) the shape of restoration process is very similar to the theoretical schedule; but 2) the overall tread has been delayed by a large margin. This means that existing I$^2$SB methods suffers from a schedule-restoration mismatching issue, \ie, there is a large discrepancy between theoretical schedule (see blue line) and practical restoration process (see blue dotted line). Such mismatching issue will results in that the theoretical nonlinear schedule is not fully leveraged on overall schedule time steps such that the inpainting performance of existing I$^2$SB method is limited.
%, which leads to the theoretical schedule are not fully leveraged on overll time steps such that the performance improvement is very limited. 

% \begin{figure}[h]
%     \centering
%     %\resizebox{0.3\linewidth}{!}{\includegraphics[]{fig/noise-schedule2.pdf}}
%     \subfigure[\textbf{high-freq pixels}]{
% 		\begin{minipage}[t]{0.31\hsize}
% 			\centering
% 			\includegraphics[width=0.98\columnwidth]{fig/restoration_speed_highfreq.pdf}
% 			\label{fig_3a}
% 		\end{minipage}%
% 	}%
%         % \quad
% 	\subfigure[\textbf{mid-freq pixels}]{
% 		\begin{minipage}[t]{0.31\hsize}
% 			\centering
% 			\includegraphics[width=0.98\columnwidth]{fig/restoration_speed_midfreq.pdf}
% 			\label{fig_3b}
% 		\end{minipage}%
% 	}%
%         % \quad
% 	\subfigure[\textbf{low-freq pixels}]{
% 		\begin{minipage}[t]{0.31\hsize}
% 			\centering
% 			\includegraphics[width=0.98\columnwidth]{fig/restoration_speed_lowfreq.pdf}
% 			\label{fig_3b}
% 		\end{minipage}%
% 	}%
%     \caption{Comparison between theoretical schedule and practical process on different frequency.}
%     \label{fig:mis-match}
% \end{figure}

% \begin{figure}
% \centering
% \includegraphics[width=0.58\hsize]{fig/restoration_speed.pdf}
% \caption{Qualitative Analysis.}
% \label{fig2}
% \end{figure}

\begin{figure}[h]
\centering
\includegraphics[width=0.98\hsize]{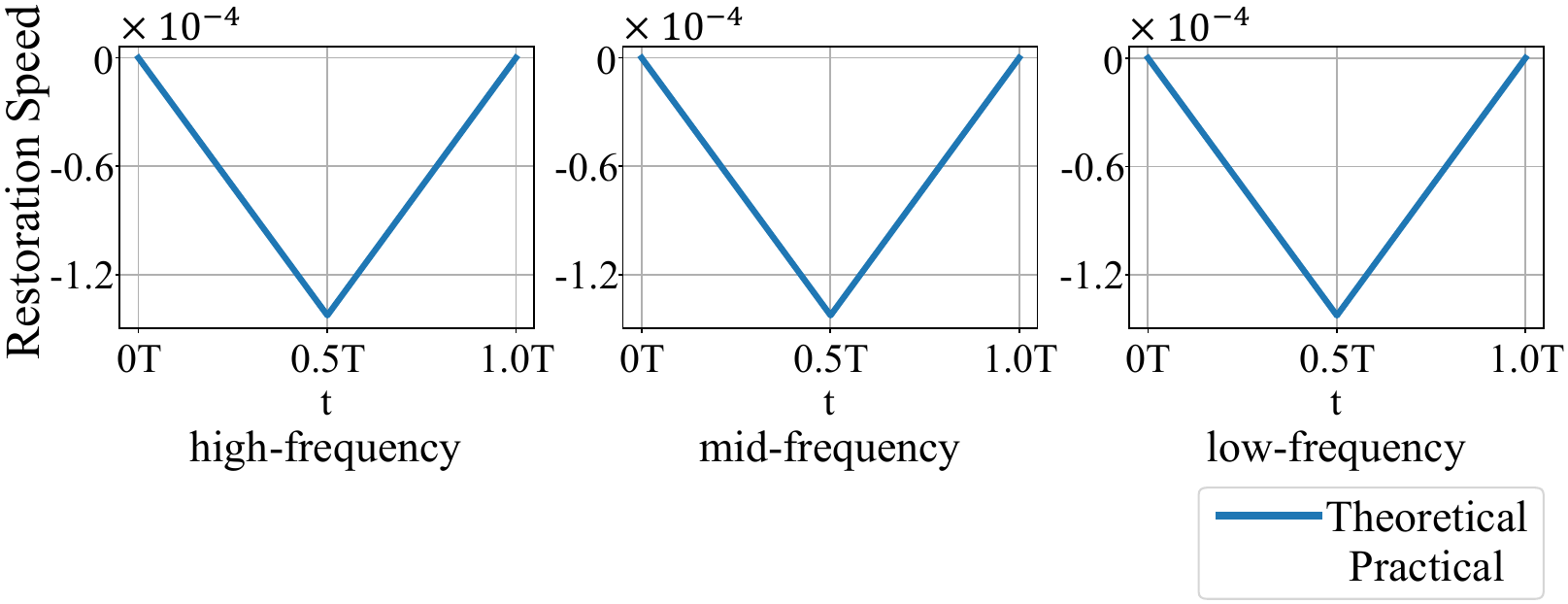}
\label{fig:mis-match}
\caption{Comparison between theoretical schedule and practical process on different frequency.}
\label{fig:mis-match}
\end{figure}

We guess the reason of causing such schedule-restoration mismatching issue is the existing I$^2$SB method sets all pixels to the same noise schedule (\ie, its assumption is the restoration process of all pixels are synchronous). However, intuitively, the generation process are not synchronous but exist a potential temporal interdependence where some high frequency pixels (\eg, outline or texture) is usually restored first and then some lower frequency pixels is repaired in later. To verify our guess, in Figure~\ref{fig:mis-match}, we evaluate the restoration process (\ie, the SSIM derivative) of CelebA-HQ test set at three type of pixels with different frequency (\ie, high-frequency pixels, middle-frequency pixels, and lower-frequency pixels), respectively. From Figure~\ref{fig:mis-match}, we can see that 1) these pixels with different frequency are indeed not restored at the same time but asynchronous; and 2) the restoration process of high-frequency pixels is very consist to the theoretical schedule, but middle- and lower-frequency pixels brings a large discrepancy which results in the overall restoration process to be delayed (\ie, causing the overall mismatching between the theoretical schedule and the practical restoration process). This fact verify our guess, \ie, the schedule-restoration mismatching issue is indeed caused by that the restoration process of all pixels are actually asynchronous but existing I$^2$SB method uses a pixel-synchronous noise schedule $\beta_t$. 

\begin{figure*}[!h]
\centering
\includegraphics[width=1.0\linewidth]{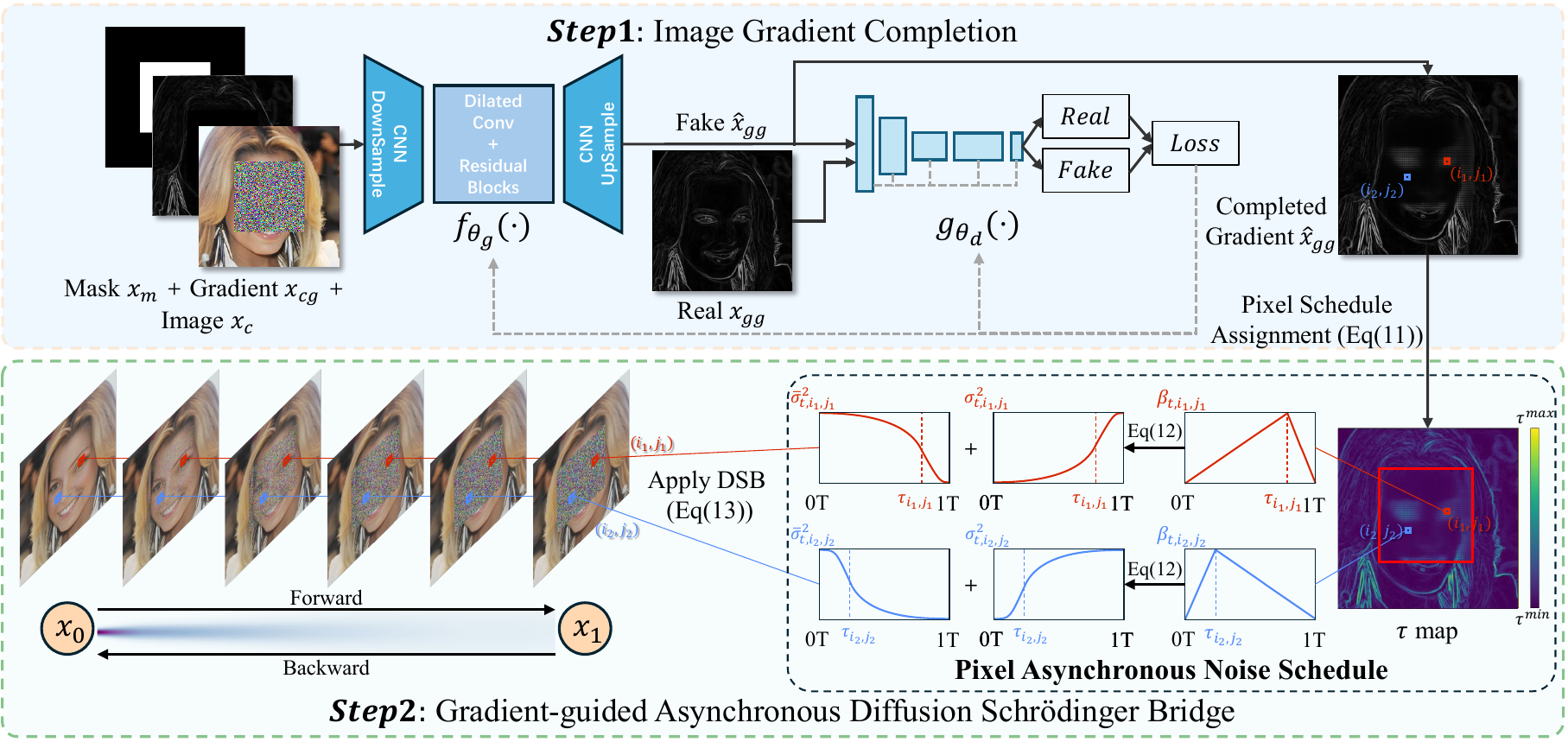} \\
\caption{The overall framework of our AsyncDSB, which consists of two steps, an image gradient completion step, and a gradient-guided asynchronous Schr\"odinger diffusion bridge step. Here, given a corrupted image, the \emph{Step 1} aims to predict the image gradient of the missing area. Then, the \emph{Step 2} accounts for assigning a pixel-asynchronous noise schedule, which is employed to make a asynchronous Schr\"odinger diffusion bridge for completing the corrupted region in origin pixel space according to the image gradient information obtained by the \emph{Step 1}.}
\label{fig:method}
\end{figure*}

\subsection{AsyncDSB Framework}
As introduced in Section~\ref{sec_3_3}, existing noise schedule methods suffers from a schedule-restoration mismatching issue, which is because that the generation process of different pixels are asynchronous but existing I$^2$SB method employs a synchronous noise schedule $\beta_t$. 
According to the above mentioned analysis, an intuitive idea is employing a non-symmetric noise schedule to match the asynchronous generation according to pixel's frequency information, \ie, the pixels with high-frequency information (\eg, contour) is preferentially scheduled and then the pixels with low-frequency information. 
To achieve this idea, we propose a schedule-Asynchronous Diffusion Schr\"odinger Bridge (AsyncDSB) framework. As shown in Figure~\ref{fig:method}, our AsyncDSB framework consists of two steps, \ie, image gradient completion and gradient-guided asynchronous Schr\"odinger diffusion bridge.

It is worth noting that the AsyncSDB framework is not limited to the image frequency guided schedule strategy. The proposed idea of asynchronous schedule actually enables any $\beta_{t,i,j}$ strategy that can match the pixel-asynchronous image generation process, which is highly potential. In this paper, we use the image frequency as guidance mainly to verify the correctness of the AsyncSDB framework.

\paragraph{\textbf{Step 1: Image Gradient Completion.}}  Intuitively, a pixel with a larger gradient has a larger frequency. To achieve the frequency prior of a corrupted image $x_c$, we first take the corrupted image $x_c$ and its mask $x_m$ and gradient map $x_{cg}$ as inputs and then design an image gradient completion module $f_{\theta_g}(\cdot)$ with parameter $\theta_g$. During inference, given a corrupted pixels $x_c$ and its mask $x_m$ and gradient map $x_{cg}$, the module $f_{\theta_g}(\cdot)$ is directly used to predict its target gradient map $\hat{x}_{gg}$. That is:
\begin{equation}
\hat{x}_{gg} =  f_{\theta_g}(x_c, x_m, x_{cg}),
\label{eq:stage1-infer}
\end{equation}
and then the gradient could be used as guidance of the asynchronous $\beta_{t,i,j}$ for each pixel.

During training, following edge prediction \cite{nazeri2019edgeconnect}, we adopt the adversarial learning manner to learn the image gradient completion module $f_{\theta_g}(\cdot)$. 
The adversarial training loss is formulated as:
\begin{equation}
\mathop{min}\limits_{\theta_g}\mathop{max}\limits_{\theta_d}L_{G} = (\lambda_{adv} \mathop{max}\limits_{\theta_d}(L_{adv})+\lambda_{FM}L_{FM}),
\label{eq:stage1-loss}
\end{equation}
where $\lambda_{adv}$ and $\lambda_{FM}$ are hyper-parameters; $L_{adv}$ and $L_{FM}$ denotes the adversarial and feature matching loss\cite{nazeri2019edgeconnect}, respectively; and $\theta_d$ denotes the parameters of discriminator $g_{\theta_d}(\cdot)$. 

\begin{table*}[t]
\footnotesize
\centering
\renewcommand\tabcolsep{4pt} % 控制列之间的间距
\resizebox{\textwidth}{!}{% 调整表格宽度适应页面
\begin{tabular}{@{}l||cccc|cccc|cccc|cccc@{}}
\toprule

\multicolumn{1}{c}{\textbf{Method}} & \multicolumn{8}{c}{\textbf{CelebA-HQ}} & \multicolumn{8}{c}{\textbf{Places2}} \\
% \cmidrule(r){1-1} \cmidrule(lr){2-9} \cmidrule(l){10-17}

 & \multicolumn{4}{c}{\textbf{FID$\downarrow$}} & \multicolumn{4}{c}{\textbf{IS$\uparrow$}} & \multicolumn{4}{c}{\textbf{FID$\downarrow$}} & \multicolumn{4}{c}{\textbf{IS$\uparrow$}} \\
\cmidrule(lr){2-5} \cmidrule(lr){6-9} \cmidrule(l){10-13} \cmidrule(l){14-17}

 & \textbf{center} & \textbf{half} & \textbf{wide} & \textbf{narrow} & \textbf{center} & \textbf{half} & \textbf{wide} & \textbf{narrow} & \textbf{center} & \textbf{half} & \textbf{wide} & \textbf{narrow} & \textbf{center} & \textbf{half} & \textbf{wide} & \textbf{narrow} \\
\cmidrule(r){1-1} \cmidrule(lr){2-2} \cmidrule(lr){3-3} \cmidrule(lr){4-4} \cmidrule(lr){5-5} \cmidrule(lr){6-6} \cmidrule(lr){7-7} \cmidrule(lr){8-8} \cmidrule(lr){9-9} \cmidrule(lr){10-10} \cmidrule(lr){11-11} \cmidrule(lr){12-12} \cmidrule(lr){13-13} \cmidrule(lr){14-14} \cmidrule(lr){15-15} \cmidrule(lr){16-16} \cmidrule(lr){17-17}
\textit{MISF} \cite{li2022misf} & 3.0 & 11.9 & 3.7 & 2.3 & 4.23 & 6.21 & 8.69 & 11.12 & 14.2 & 26.4 & 7.7 & 4.5 & 61.3 & 37.3 & 65.8 & 77.3 \\
% \textit{PUT} \cite{yan2019PENnet} & - & - & - & - & - & - & - & - & - & - & - & - & - & - & - & - \\
\textit{ZITS} \cite{dong2022incremental} & 4.6 & 10.9 & 5.4 & 3.2 & 5.28 & 7.24 & 8.63 & 10.32 & \underline{8.3} & 11.7 & \underline{5.4} & 3.2 & \underline{64.9} & 51.6 & 69.8 & 79.6 \\ \hline
% \textit{LantentPaint} \cite{cao2022learning} & CVPR'24 & - & - & - & - & - & - & - & - & - & - & 7.1 & 6.8 \\
\textit{StrDiffusion} \cite{liu2024structure} & 6.7 & 17.3 & 8.3 & 3.7 & 5.46 & 7.08 & 8.83 & 11.10 & 16.0 & 29.6 & 8.5 & 3.7 & 57.1 & 33.9 & 63.9 & 78.3 \\
\textit{RePaint} \cite{lugmayr2022repaint} & 9.0 & 12.1 & 7.1 & 7.3 & 6.40 & 8.25 & \underline{9.61} & \underline{11.31} & 9.0 & 13.0 & 5.5 & 3.4 & 64.2 & 49.12 & \textbf{71.1} & 79.6 \\
\textit{DiffIR$_{S2}$} \cite{xia2023diffir} & 2.2 & 9.7 & 3.4 & 3.7 & 5.86 & 8.34 & 9.60 & 11.27 & 8.8 & 11.1 & 5.7 & 3.7 & 64.9 & \underline{51.6} & \underline{70.2} & \underline{80.0} \\ \hline
\textit{BBDM} \cite{li2023bbdm} & 6.9 & 8.5 & 7.4 & 7.5 & 6.00 & 6.43 & 6.46 & 6.42 & 28.5 & 32.6 & 30.6 & 27.1 & 31.8 & 26.1 & 26.8 & 33.9 \\
\textit{I$^{2}$SB (baseline)} \cite{liu2023i2sb} & \underline{2.2} & \underline{5.4} & \underline{2.9} & \underline{1.7} & \underline{6.71} & \underline{8.35} & 9.50 & 11.30 & 8.6 & \underline{10.8} & 5.6 & \underline{3.1} & 64.6 & 43.3 & 69.8 & 79.7 \\
\midrule
%\textbf{AsyncDSB (A)} & \textbf{1.9} & \textbf{5.2} & \textbf{2.9} & \textbf{1.7} & \textbf{6.73} & \textbf{8.38} & \textbf{9.55} & \textbf{11.37} & \textbf{8.3} & \textbf{10.5} & \textbf{5.3} & \textbf{3.0} & \textbf{65.1} & \textbf{52.3} & 69.9 & \textbf{80.0} \\
% sum(FID)=38.8 sum(IS)=303.33
\textbf{AsyncDSB} & \textbf{1.9}$_{\blacktriangle 14\%}$ & \textbf{5.2}$_{\blacktriangle 4\%}$ & \textbf{2.8}$_{\blacktriangle 3\%}$ & \textbf{1.6}$_{\blacktriangle 6\%}$ & \textbf{6.74}$_{\blacktriangle 4\text{\textperthousand}}$ & \textbf{8.38}$_{\blacktriangle 4\text{\textperthousand}}$ & \textbf{9.62}$_{\blacktriangle 12\text{\textperthousand}}$ & \textbf{11.41}$_{\blacktriangle 9\text{\textperthousand}}$ & \textbf{8.3}$_{\blacktriangle 4\%}$ & \textbf{10.5}$_{\blacktriangle 3\%}$ & \textbf{5.2}$_{\blacktriangle 7\%}$ & \textbf{3.0}$_{\blacktriangle 3\%}$ & \textbf{65.1}$_{\blacktriangle 8\text{\textperthousand}}$ & \textbf{52.3}$_{\blacktriangle 14\text{\textperthousand}}$ & 69.9$_{\blacktriangle 1\text{\textperthousand}}$ & \textbf{80.0}$_{\blacktriangle 1\text{\textperthousand}}$ \\
% sum(FID)=38.5 sum(IS)=303.45
\bottomrule
\end{tabular}
} % end of resizebox
\caption{Quantitative results on CelebA-HQ and Places2. $\downarrow$: Lower is better for FID; $\uparrow$: Higher is better for IS. The best and second-best results are marked with \textbf{boldface} and \underline{underline}, respectively.}
\label{table:compare2}
\end{table*}
\begin{figure*}[h!]
    \centering
    %\resizebox{0.3\linewidth}{!}{\includegraphics[]{fig/noise-schedule2.pdf}}
    \subfigure[\textbf{Regular masks (\ie, center and half)}]{
		\begin{minipage}[t]{0.50\textwidth}
			\centering
			\includegraphics[width=0.998\columnwidth]{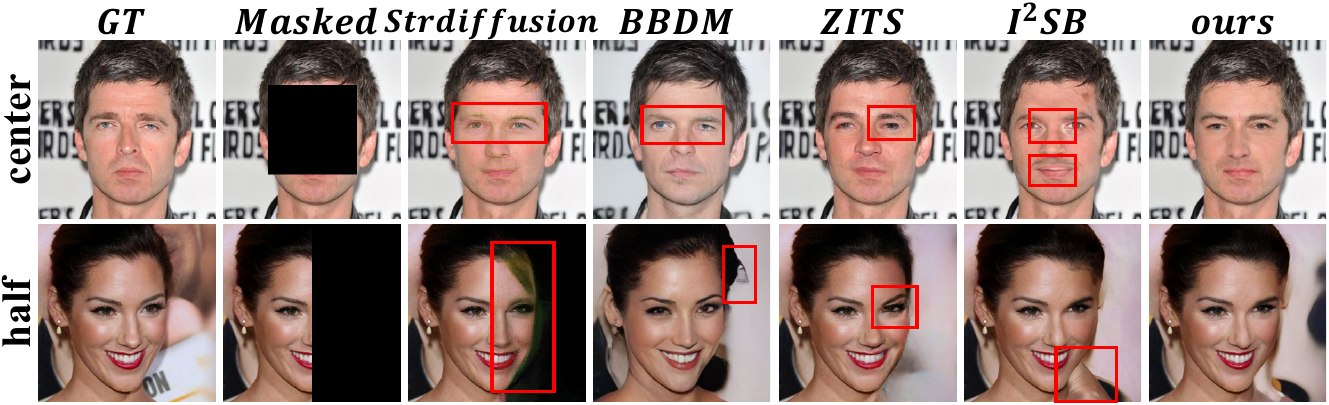}
			\label{fig_3a}
		\end{minipage}%
	}%
	\subfigure[\textbf{Irregular masks (\ie, wide and narrow)}]{
		\begin{minipage}[t]{0.50\textwidth}
			\centering
			\includegraphics[width=0.998\columnwidth]{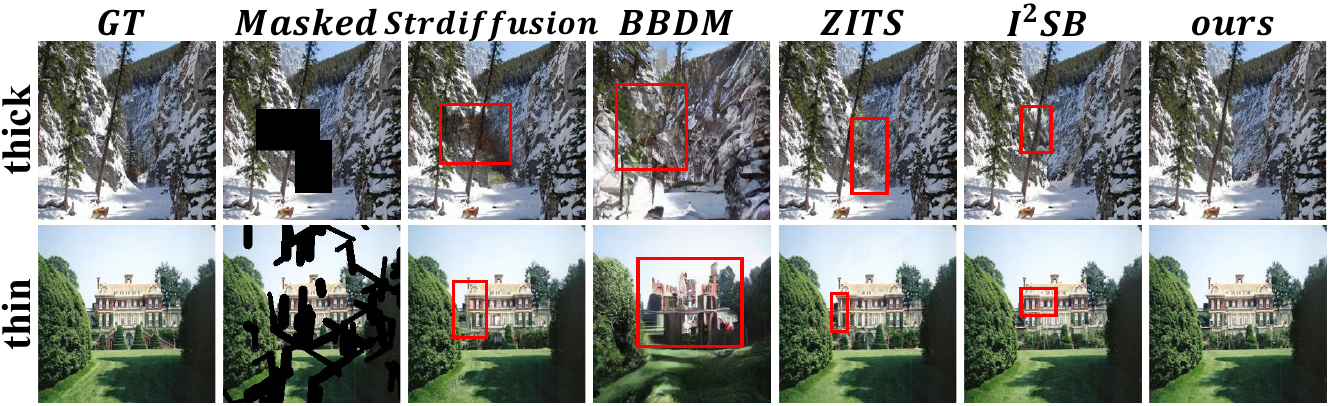}
			\label{fig_3b}
		\end{minipage}%
	}%
    \caption{Qualitative results on CelebA-HQ and Places2 under regular and irregular masks.}
    \label{fig5}
\end{figure*}

\paragraph{\textbf{Step 2: Gradient-guided Asynchronous Diffusion Schr\"odinger Bridge.}}
As analyzed in Section\ref{sec_3_3}, high-frequency pixels should restore earlier than pixels with low frequency. To this end, we introduce a new variable $\tau_{i,j} \in [0T, 1.0T]$ and transform the traditional noise schedule $\beta_t$ into an asymmetric and pixel-wise noise schedule form, \ie, $\beta_{t,i,j}$ (see bottom right part of Figure~\ref{fig:method} for its details). The advantage of such design is that the schedule sequence between pixels can be freely adjusted by assigning different variable $\tau_{i,j}$. Intuitively, the pixel with high frequency should assign larger $\tau_{i,j}$ than pixel with low frequency since the pixel with high frequency are priority restored. To achieve this idea, we design a simple yet effective pixel-wise schedule assignment strategy. That is,
\begin{equation}
\resizebox{.89\hsize}{!}{$
\tau_{i,j} = \frac{gauss(\hat{x}^{gg}_{i,j})-min(gauss(\hat{x}_{gg}))}{max(gauss(\hat{x}_{gg}))-min(gauss(\hat{x}_{gg}))}(\tau^{max}-\tau^{min}) + \tau^{min}
\label{eq:stage2-tau}
$}
\end{equation}
where $\tau^{max}$ and $\tau^{min}$ are two hyper-parameters, which controls the asynchronous range of pixel schedule; and $gauss()$ is a Gaussian filter, which aims to achieve a flat noise schedule for all pixels. 

After assigning noise schedule for each pixel, we follow Eq~\eqref{eq3} to calculate two variances $\sigma_{t,i,j}$ and $\overline{\sigma}_{t,i,j}$ for each pixel by accumulating from either sides along a noise schedule $\beta_{t,i,j}$. That is,
\begin{equation}
\sigma_{t,i,j}^2 = \int_{0}^{t}\beta_{\tau,i,j} d\tau,\ \overline{\sigma}_{t,i,j}^2 = \int_{t}^{1}\beta_{\tau,i,j} d\tau.
\label{eq12}
\end{equation}
Finally, during inference, we directly use the pixel form of Eq~\eqref{eq:ddpm} to restore images. That is,
\begin{equation}
\resizebox{.88\hsize}{!}{$
\begin{aligned}
&q(x_{t-\frac{1}{T},i,j}|x_{t,i,j}) \\ 
&= \mathcal{N} \left (x_{t,i,j}-\int_{t-\frac{1}{T}}^{t}\left [ -\beta_{t,i,j} \nabla \log \hat{\Psi}(x_{t,i,j}, t) \right ]dt, \int_{t-\frac{1}{T}}^{t} \beta_{t,i,j} dt \right ),
\end{aligned}
$}
% \label{eq:ddpm}
\end{equation} where $\nabla log \hat{\Psi}(x_{t,i,j}, t) = \frac{s_{\theta}(x_{t,i,j}, t)}{\sigma_{t}}$. 
% Please see Section~\ref{sec_3_5} for more training and inference details.
Please see Appendix for more training and inference details.

\section{Experiments}
% 有没有可能AsyncDSB能用更少的setp保持好的效果，或者step更少但效果比其他方法还好，这样也可以加实验？如果Async的schedule比较合理，那它应该能更早生成图像。比如都用三个step，base方法应该在第二个step能走一大步，其他step走的小，而Async可以走三个较大的步，比base的要远。如果是这样那可以画一个示意图在第二页
\subsection{Experimental Settings}
\noindent \textbf{Dataset.}
We validate our method and various baselines over two typical dataset: 
\textbf{1) CelebA-HQ} is a face dataset. We split this dataset by following standard dataset split \cite{liu2015faceattributes}. 
\textbf{2) Places2} is a scene image dataset, which split into 1.8 million training images and 36500 validation images, from which we randomly select 6000 images as test. Following RePaint \cite{lugmayr2022repaint}, we use four types of masks (\ie, center, half, wide and narrow) to conduct experiments where all images are set to 256x256 resolution. %Among them, center and half are fixed masks at the center and right half, while wide and narrow are random masks generated according to the settings of RePaint. Note that all datasets we use are at 256x256 resolution.

\noindent \textbf{Evaluation Metrics.}
Following \cite{liu2023i2sb}, we employ two widely used metrics, Fréchet Inception Distance (FID) and Inception Score (IS), to evaluate the quality and diversity of the generated images. %FID measures the similarity between the generated and real image distributions, while IS assesses the quality and diversity of the generated samples. Lower FID and higher IS scores indicate better performance of the model in generating realistic and diverse images.

% \noindent \textbf{Implementation Details.}
% For fairness, we use the same settings as I$^2$SB to design and train our UNet. We set the time step $T$ to 1000, the batch size to 8, and sobel operator to obtain image gradients. 
%All experiments were implemented using PyTorch and run on an NVIDIA RTX A6000. 

% \begin{figure}
%     \centering
%     %\resizebox{0.3\linewidth}{!}{\includegraphics[]{fig/noise-schedule2.pdf}}
%         \centering
%         \includegraphics[width=0.98\columnwidth]{fig/visual-compare.pdf}
%         \caption{Qualitative results on CelebA-HQ and Places2. Here, the red-color boxes highlight details.}
%     \label{fig:visual-compare}
% \end{figure}

\subsection{Results and Discussion}
We select seven latest methods: MISF \cite{li2022misf}, ZITS \cite{dong2022incremental}, StrDiffusion \cite{liu2024structure}, RePaint \cite{lugmayr2022repaint}, DiffIR$_{S2}$ \cite{xia2023diffir}, BBDM \cite{li2023bbdm}, and I$^{2}$SB \cite{liu2023i2sb} as our baseline. The results are shown in Table~\ref{table:compare2} and Figure~\ref{fig5}.

\noindent\textbf{Quantitative Evaluation.} In Table \ref{table:compare2}, we  evaluate all methods on CelebA-HQ and Places2 under varied mask settings. We can see that 1) our AsyncDSB achieves the best performance, especially on FID, which verifies our superiority; 2) our AsyncDSB exceeds the key competitor (\ie, I$^2$SB) around 3\% $\sim$ 14\%. This means that employing a pixel-asynchronous noise schedule is very beneficial for I$^2$SB, which verifies the effectiveness of the proposed AsyncDSB; 3) the performance improvement is more significant on center\&half masks than wide\&narrow, which may be because the missing of high-frequency information (\eg, outline) is even worse on center\&half masks.
% Specifically, compared with I$^2$SB, our method better matches the schedule and resotrement process by introducing a pixel asynchronous noise schedule from predicted gradient information as priors.
% It‘s worth noting that our method also beats other image inpainting methods that use diffusion models(\ie, Strdiffusion, RePaint, DiffIR$_{S2}$) or diffusion bridge models (\ie, BBDM).
%Specifically, under the center mask setting on CelebA-HQ, our method outperforms the second-best method I$^2$SB by 13.6\% in terms of FID. For the half mask on CelebA-HQ, our method surpasses I$^2$SB by 3.7\% on FID. On the Places2 dataset, our method achieves a 3.5\% improvement on FID under the center mask setting compared to the second-best method ZITS. For the half mask on Places2, our method outperforms I$^2$SB by 2.8\% on FID. These results demonstrate the effectiveness of our AsynDSB in inpainting high-quality and diverse images. 

\noindent\textbf{Qualitative Evaluation.}
In Figure.~\ref{fig5}, we show some visual results of our AsyncDSB and some representative methods. It can be observed that our AsyncDSB generates images with finer details and more natural textures that are well blended with the context. This further verifies the effectiveness.

\subsection{Ablation Study}
\label{sec:ablation}
\begin{figure}
\centering
\includegraphics[width=.98\hsize]{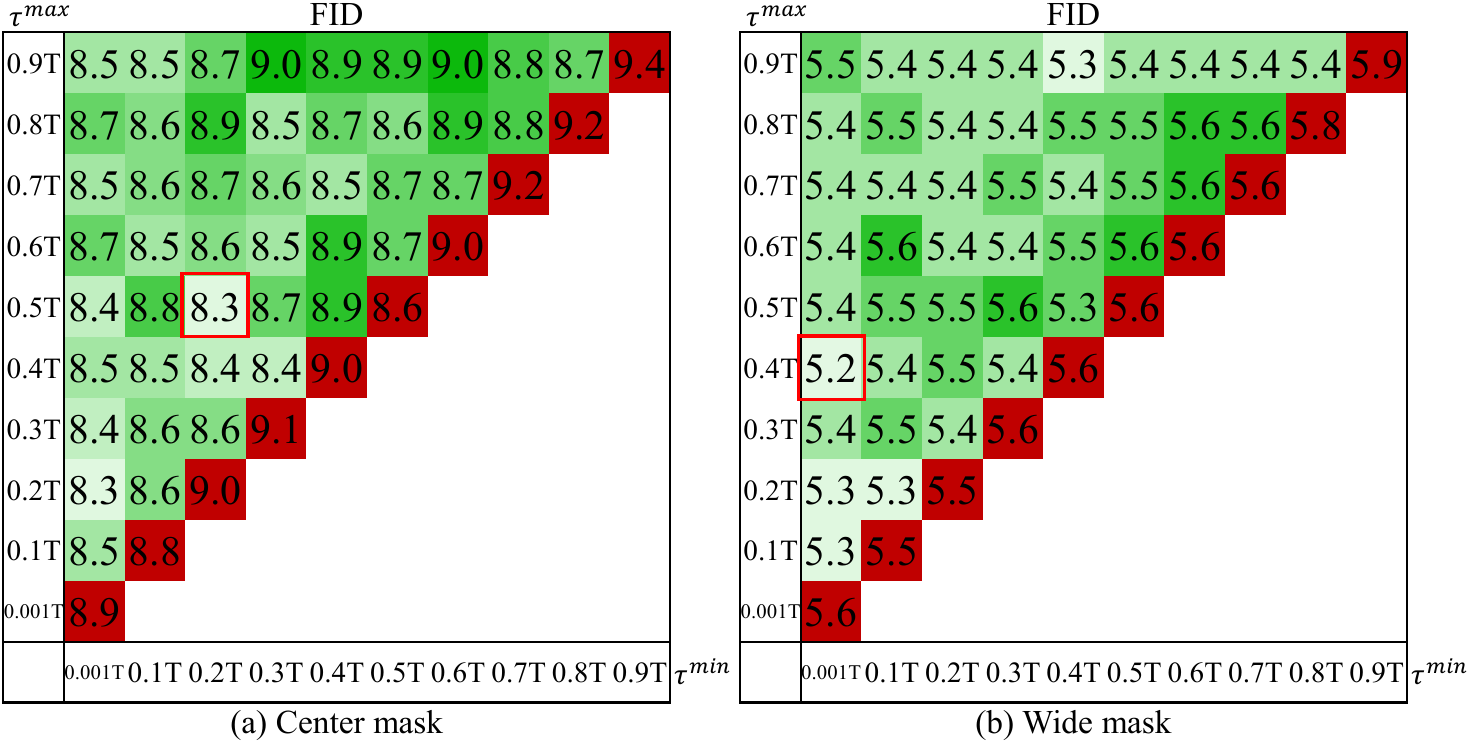}
\caption{Analysis of $\tau^{min}$ and $\tau^{max}$ on Places2.}
\vspace{-10pt}
\label{fig:ras_example}
\end{figure}

\noindent \textbf{Hyper-parameters ($\tau^{min}$ and $\tau^{max}$) Setting}
In Figure~\ref{fig:ras_example}, we conduct a parameter analysis on Places 2 from $\tau^{min}=0.001T$ to $\tau^{min}=0.9T$ and $\tau^{max}=0.001T$ to $\tau^{min}=0.9T$. In particular, the blank value in Figure~\ref{fig:ras_example} is not valid since two parameters must satisfy $\tau^{min}\leq\tau^{max}$. From results, we can see that 1) compared with the setting of $\tau^{min}==\tau^{max}$ (\ie, pixel-synchronous schedule), our AsyncDSB achieves more superior performance when $\tau^{min}$ is set to a smaller value, which further verifies the effectiveness of our pixel-asynchronous schedule; 2) our AsyncDSB achieves best performance when we set $(\tau^{min}, \tau^{max})$ to (0.2$T$, 0.5$T$) for the regular masks like center and (0.001$T$, 0.4$T$) for the irregular masks like wide. We guess that this phenomenon may be caused by the loss of different frequencies caused under different masks.

\begin{figure}
\centering
\includegraphics[width=0.98\hsize]{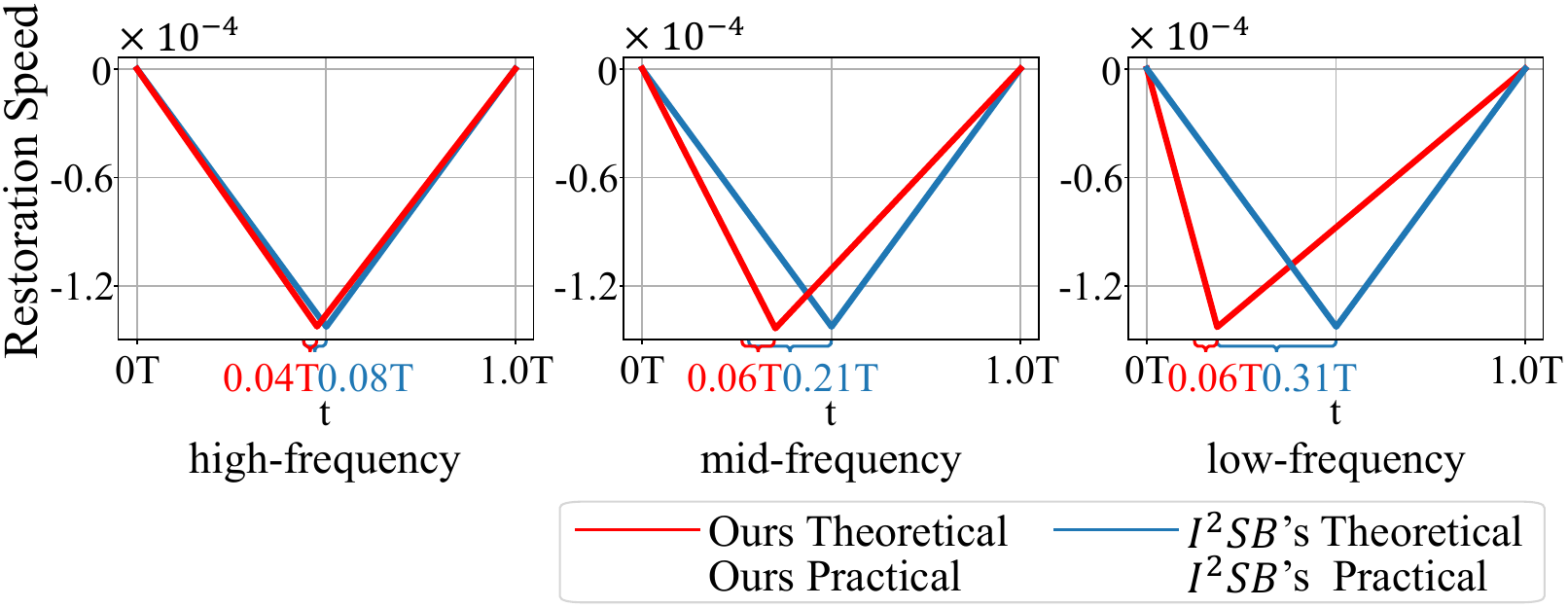}
\label{fig:match1}
\captionof{figure}{Comparison between theoretical schedule and practical process on different frequency.}
\vspace{-10pt}
\label{fig:match1}
\end{figure}

% \noindent \textbf{What's the connection between the hyper-parameter and the mask ratio?}

\noindent \textbf{Hyper-parameter ($\tau^{min}$ and $\tau^{max}$) and Mask Ratio}
Intuitively, a large mask ratio is easier to cause the loss of high-frequency information since most pixels are low-frequency. To this end, in Table~\ref{tab:para-ratio}, we report the optimal parameters for different mask ratio setting. We find that 1) large mask ratios tend to the large $\tau_{min}$ and $\tau_{max}$, \ie, tend to set earlier schedule for each pixel; 2) the performance achieve consistent improvement. This means exploring a mask-and-sample-specific pixel-asynchronous noise schedule is promising for enhancing inpainting performance. We will explore it in future work.

\begin{table}[t]
\centering
\scalebox{0.8}{
\begin{tabular}{lccc}
\toprule
\textbf{Mask Ratios} & \multicolumn{2}{c}{\textbf{hyper-parameters}} & \textbf{Improved FID} \\
 & \textbf{$\tau_{min}$} & \textbf{$\tau_{max}$} \\
\midrule
$1\%\sim 10\%$ &  0.001$T$ & 0.9$T$ & 7.6$\rightarrow$7.5\\
$11\%\sim 20\%$ & 0.2$T$ & 0.9$T$ & 13.1$\rightarrow$13.0\\
$21\%\sim 30\%$ & 0.4$T$ & 0.9$T$ & 17.6$\rightarrow$17.5\\
$31\%\sim 40\%$ & 0.7$T$ & 0.9$T$ & 26.2$\rightarrow$26.1\\
\bottomrule
\end{tabular}}
\caption{Analysis of mask ratios on Places2.}
\vspace{-10pt}
\label{tab:para-ratio}
\end{table}

\begin{table}[h!]
\centering
\scalebox{0.8}{
\begin{tabular}{lcccc }
\toprule
\textbf{Settings} & \multicolumn{4}{c}{\textbf{FID$\downarrow$}} \\
 & \textbf{center} & \textbf{half} & \textbf{wide} & \textbf{narrow} \\
\midrule
%Distance Guided & 2.1 & 6.5 & 3.0 & 1.8 \\
I$^2$SB+gradient & 2.0 & 5.8 & 3.0 & 1.7 \\
Our AysncDSB & 1.9 & 5.2 & 2.8 & 1.6 \\
% w/ gaussian-11 & 2.0 & 5.7 & 3.0 & 1.7 \\
% w/ gaussian-31 & 1.9 & 5.2 & 2.8 & 1.6\\
\bottomrule
\end{tabular}}
\caption{Analysis on asynchronous schedule.}
\vspace{-10pt}
\label{tab:struct_dist1}
\end{table}

\noindent \textbf{Pixel-asynchronous Schedule  v.s. Synchronous Schedule}
In Table~\ref{tab:struct_dist1}, we add the image gradient into I$^2$SB (\ie, regarding as UNet condition) as baseline, which can be regarded as a gradient-guided pixel-synchronous I$^2$SB and report its performance on CelebA-HQ. We find that our AsyncDSB exceeds the baseline by a large margin, which means that our pixel-asynchronous schedule is effective. 

\begin{figure}[h!]
\centering
\includegraphics[width=0.98\hsize]{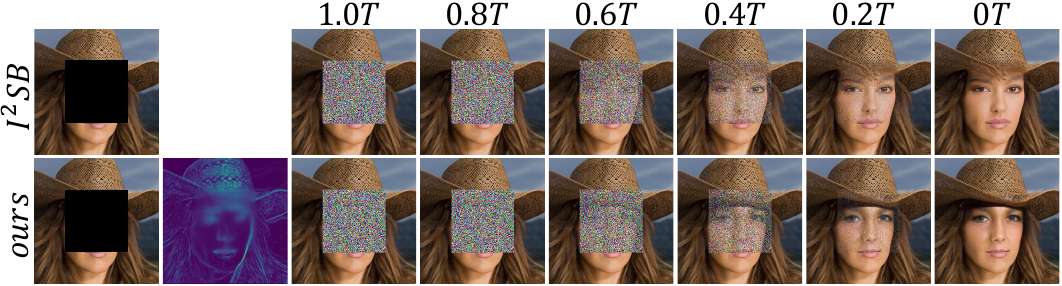} \\
\caption{Sampling process of our AsyncDSB.}
\vspace{-10pt}
\label{fig:visual-ablation}
\end{figure}

\noindent \textbf{Can AsyncDSB solve the schedule-restoration mismatch issue?} In Figure~\ref{fig:match1}, we compare the restoration process (\ie, the SSIM derivative) between I$^2$SB and our AsyncDSB on CelebA-HQ test set at different frequency pixels. From results, we can see that the schedule-restoration mismatch issue can be significantly alleviated after applying our AsyncDSB, which verifies our effectiveness.

% \begin{figure}[h]
%     \centering
%     %\resizebox{0.3\linewidth}{!}{\includegraphics[]{fig/noise-schedule2.pdf}}
%     \subfigure[\textbf{high-frequency}]{
% 		\begin{minipage}[t]{0.45\hsize}
% 			\centering
% 			\includegraphics[width=0.98\columnwidth]{fig/match-speed-high.pdf}   
% 			\label{fig_8a}
% 		\end{minipage}%
% 	}%
%         % \quad
% 	\subfigure[\textbf{mid-frequency}]{
% 		\begin{minipage}[t]{0.45\hsize}
% 			\centering
% 			\includegraphics[width=0.98\columnwidth]{fig/match-speed-mid.pdf}
% 			\label{fig_8b}
% 		\end{minipage}%
% 	}%
%         \quad
% 	\subfigure[\textbf{low-frequency}]{
% 		\begin{minipage}[t]{0.45\hsize}
% 			\centering
% 			\includegraphics[width=0.98\columnwidth]{fig/match-speed-low.pdf}
% 			\label{fig_8b}
% 		\end{minipage}%
% 	}%
%     \caption{Analysis of schedule-restoration mismatch issue on Places 2 under center mask.}
%     \label{fig:match1}
% \end{figure}

% \noindent \textbf{How does our AsyncDSB work?}

\noindent \textbf{How does our AsyncDSB work?}
In Figure~\ref{fig:visual-ablation}, we random select two test image from CelebA-HQ and visualize its completed image gradient and restoration process. From Figure~\ref{fig:visual-ablation}, we can see that 1) our AsyncDSB well competed the missing gradient map and 2) compared with pixel-synchronous schedule (\ie, I$^2$SB), in our AsyncDSB, some high-frequency content (see the outline of hat at $t=0.66T$) indeed do recover earlier than low-frequency details (see the face color at $t=0.33T$).

\section{Conclusion}
In this paper, we find that existing diffusion Schr\"odinger bridge method suffers from a schedule-restoration mismatch issue, which is caused by setting all pixels to a synchronous noise schedule. To this end, we propose a schedule-asynchronous diffusion Schr\"odinger bridge for image inpainting. Its advantage is resorting to the asynchronous schedule at pixels, the temporal interdependence between pixels can be fully characterized for high-quality image inpainting. Experimental results show that our method
achieves superior performance over previous state-of-the-art methods.

\section{Acknowledgments}
This work was supported by the Shenzhen Peacock Program under Grant No. ZX20230597, Stabilization Support Program (ZX20240460), NSFC under Grant No. 62272130 and Grant No. 62376072, and the Shenzhen Science and Technology Program under Grant No. KCXFZ20211020163403005.

\bibliography{aaai25}

\section{Reproducibility Checklist}

This paper:

\begin{itemize}
\item Includes a conceptual outline and/or pseudocode description of AI methods introduced (yes)
\item Clearly delineates statements that are opinions, hypothesis, and speculation from objective facts and results (yes)
\item Provides well marked pedagogical references for less-familiare readers to gain background necessary to replicate the paper (yes)
\end{itemize}
Does this paper make theoretical contributions? (yes)

If yes, please complete the list below.

\begin{itemize}
\item All assumptions and restrictions are stated clearly and formally. (yes)
\item All novel claims are stated formally (e.g., in theorem statements). (yes)
\item Proofs of all novel claims are included. (yes)
\item Proof sketches or intuitions are given for complex and/or novel results. (yes)
\item Appropriate citations to theoretical tools used are given. (yes)
\item All theoretical claims are demonstrated empirically to hold. (yes)
\item All experimental code used to eliminate or disprove claims is included. (yes)
\end{itemize}

Does this paper rely on one or more datasets? (yes)

If yes, please complete the list below.
\begin{itemize}
\item A motivation is given for why the experiments are conducted on the selected datasets (yes)
\item All novel datasets introduced in this paper are included in a data appendix. (yes)
\item All novel datasets introduced in this paper will be made publicly available upon publication of the paper with a license that allows free usage for research purposes. (yes)
\item All datasets drawn from the existing literature (potentially including authors’ own previously published work) are accompanied by appropriate citations. (yes)
\item All datasets drawn from the existing literature (potentially including authors’ own previously published work) are publicly available. (yes)
\item All datasets that are not publicly available are described in detail, with explanation why publicly available alternatives are not scientifically satisficing. (yes)
\end{itemize}

Does this paper include computational experiments? (yes)

If yes, please complete the list below.
\begin{itemize}
\item Any code required for pre-processing data is included in the appendix. (yes).
\item All source code required for conducting and analyzing the experiments is included in a code appendix. (yes)
\item All source code required for conducting and analyzing the experiments will be made publicly available upon publication of the paper with a license that allows free usage for research purposes. (yes)
\item All source code implementing new methods have comments detailing the implementation, with references to the paper where each step comes from (yes)
\item If an algorithm depends on randomness, then the method used for setting seeds is described in a way sufficient to allow replication of results. (yes)
\item This paper specifies the computing infrastructure used for running experiments (hardware and software), including GPU/CPU models; amount of memory; operating system; names and versions of relevant software libraries and frameworks. (yes)
\item This paper formally describes evaluation metrics used and explains the motivation for choosing these metrics. (yes)
\item This paper states the number of algorithm runs used to compute each reported result. (yes)
\item Analysis of experiments goes beyond single-dimensional summaries of performance (e.g., average; median) to include measures of variation, confidence, or other distributional information. (yes)
\item The significance of any improvement or decrease in performance is judged using appropriate statistical tests (e.g., Wilcoxon signed-rank). (no)
\item This paper lists all final (hyper-)parameters used for each model/algorithm in the paper’s experiments. (yes)
\item This paper states the number and range of values tried per (hyper-) parameter during development of the paper, along with the criterion used for selecting the final parameter setting. (yes)
\end{itemize}

\end{document}